\documentclass[10pt,twocolumn,letterpaper]{article}

\usepackage[pagenumbers]{cvpr} % To force page numbers, e.g. for an arXiv version
\newcommand\blfootnote[1]{%
  \begingroup
  \renewcommand\thefootnote{}\footnote{#1}%
  \addtocounter{footnote}{-1}%
  \endgroup
}

\definecolor{figgray}{RGB}{51, 51, 51}

\usepackage{xspace}
\newcommand{\method}{Reshoot-Anything\xspace}

\newcommand{\recammaster}{$\text{ReCamMaster}$}
\newcommand{\trajectorycrafter}{$\text{TrajectoryCrafter}$}
\newcommand{\exfd}{$\text{EX-4D}$}

\definecolor{cvprblue}{rgb}{0.21,0.49,0.74}
\usepackage[pagebackref,breaklinks,colorlinks,allcolors=cvprblue]{hyperref}

\usepackage{enumitem}
\usepackage{multirow}
\usepackage{makecell}
\usepackage{bm}
\usepackage{xcolor}

\def\paperID{10} % *** Enter the Paper ID here
\def\confName{CVPR}
\def\confYear{2026}

\title{\method{}: A Self-Supervised Model for In-the-Wild Video Reshooting  \vspace{-0.3cm}}

\author{Avinash Paliwal$^{*}$ \quad Adithya Iyer$^{*}$ \quad Shivin Yadav \quad Muhammad Ali Afridi \quad Midhun Harikumar\\
\small \href{https://adithyaiyer1999.github.io/reshoot-anything/}{https://adithyaiyer1999.github.io/reshoot-anything} \\
\small Morphic Inc.}
\begin{document}
\twocolumn[{
\renewcommand\twocolumn[1][]{#1}
\maketitle
\footnotetext[1]{$^{*}$Equal contribution.}
\vspace{-0.29in}
\begin{center}
  \centering
  \vspace{-0.2in}
  \includegraphics[width=0.8\linewidth]{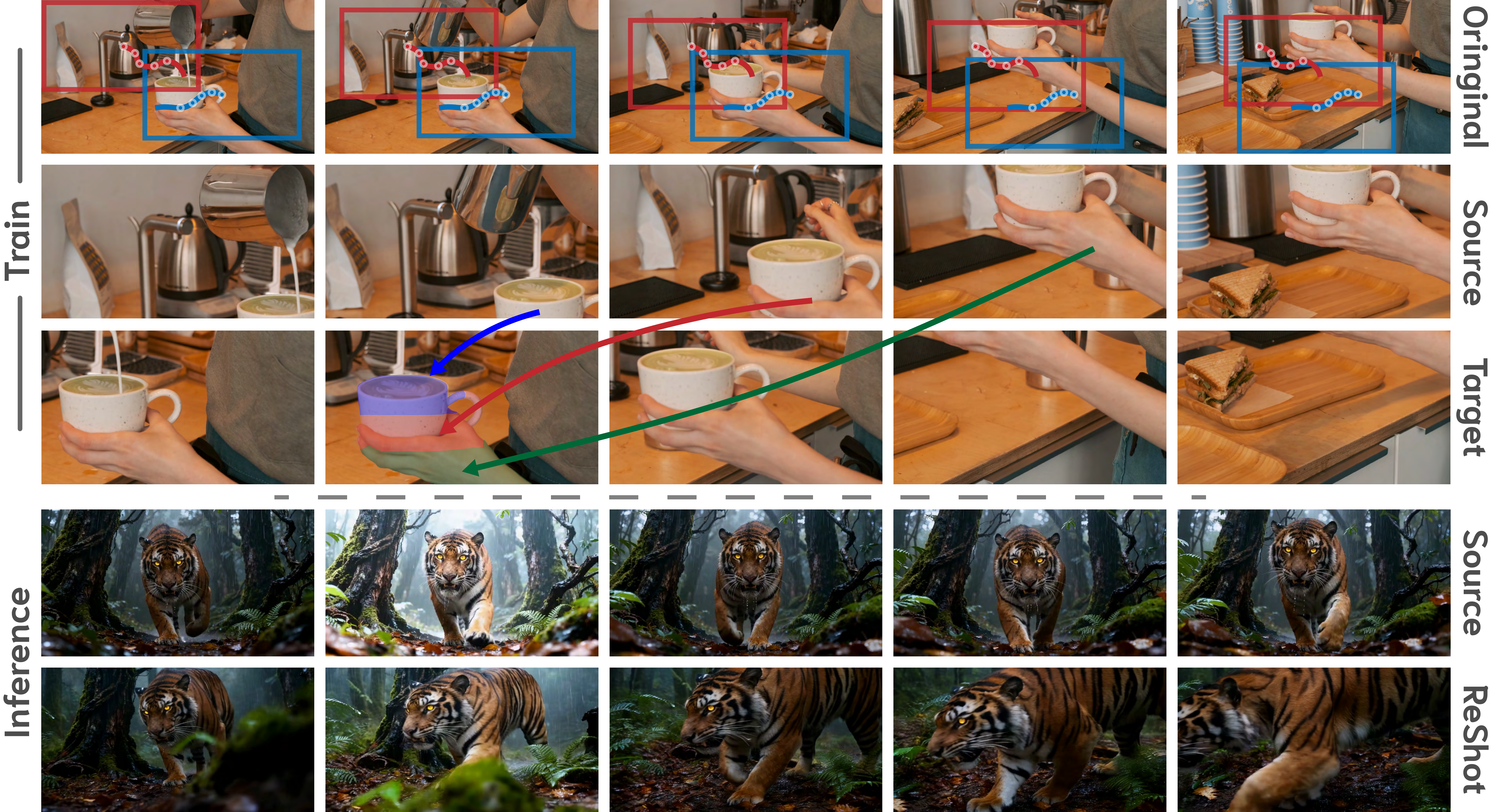}
  \vspace{-0.1in}
  % \vspace{-10pt}
    \captionof{figure}{\textbf{Reshooting Video from Novel Viewpoints.} We introduce a self-supervised framework for dynamic video reshooting trained on \textit{monocular videos}. \textbf{Top:} We generate pseudo multi-view triplets by sampling distinct crop trajectories (\textcolor{red}{red} for source, \textcolor{blue}{blue} for target) from a single video. As a result, the target view often requires regions occluded in the corresponding source frame. To reconstruct the target object (e.g., the cup and hand), the model must route missing textures from multiple different source frames (indicated \textcolor{blue}{by} \textcolor{red}{colored} \textcolor{teal}{arrows} and matching overlays). This spatial bottleneck compels the model to learn implicit 4D spatiotemporal structures. \textbf{Bottom:} At inference, our model leverages these learned 4D priors to reshoot complex dynamic monocular videos under novel camera trajectories.}
  \label{fig:teaser} 
\end{center}
}
]
\blfootnote{$^{*}$Equal contribution.}   % <-- add this line

\begin{abstract}
Precise camera control for reshooting dynamic videos is bottlenecked by the severe scarcity of paired multi-view data for non-rigid scenes. We overcome this limitation with a highly scalable self-supervised framework capable of leveraging internet-scale monocular videos. Our core contribution is the generation of pseudo multi-view training triplets, consisting of a source video, a geometric anchor, and a target video. We achieve this by extracting distinct smooth random-walk crop trajectories from a single input video to serve as the source and target views. The anchor is synthetically generated by forward-warping the first frame of the source with a dense tracking field, which effectively simulates the distorted point-cloud inputs expected at inference. Because our independent cropping strategy introduces spatial misalignment and artificial occlusions, the model cannot simply copy information from the current source frame. Instead, it is forced to implicitly learn 4D spatiotemporal structures by actively routing and re-projecting missing high-fidelity textures across distinct times and viewpoints from the source video to reconstruct the target. At inference, our minimally adapted diffusion transformer utilizes a 4D point-cloud derived anchor to achieve state-of-the-art temporal consistency, robust camera control, and high-fidelity novel view synthesis on complex dynamic scenes.
\end{abstract}    
\vspace{-0.1in}
\section{Introduction}
\label{sec:intro}

% \figgray{\bm{$z_s$}}

\begin{figure}
    \centering
    \includegraphics[width=\linewidth]{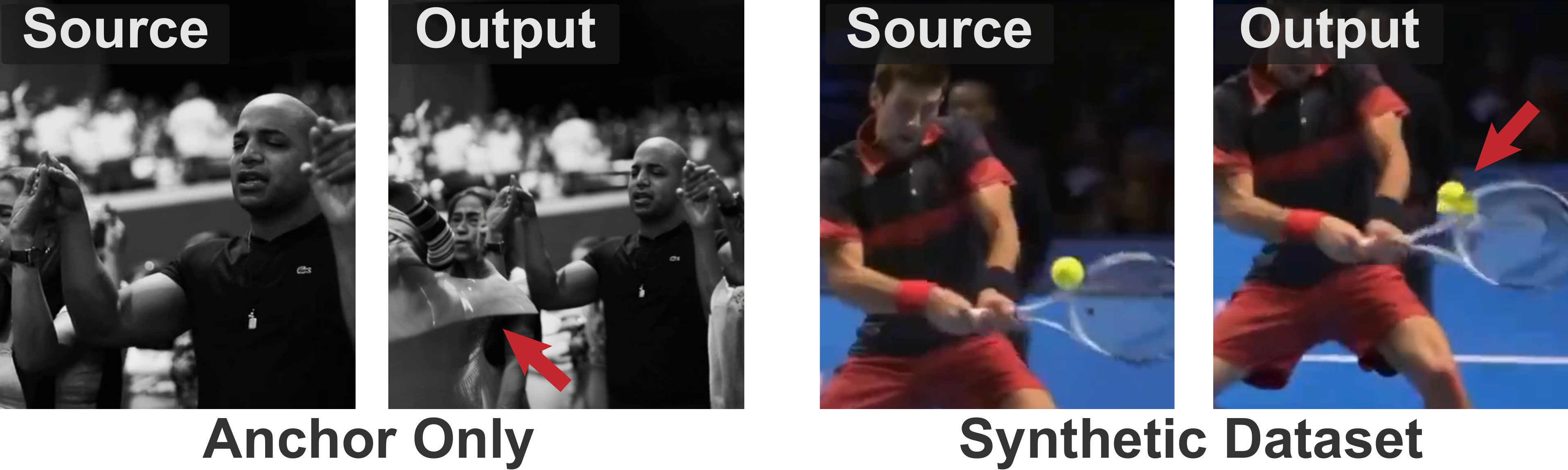}
    \vspace{-0.25in}
    \caption{A comparison of video reshooting challenges. (a) Anchor-only methods~\cite{ex4d} are prone to artifacts when the anchor video $V_a$ is of low quality. (b) Synthetic data trained models~\cite{bai2025recammaster} can fail to generalize and may produce artifacts on unseen interactions, like a tennis ball being hit.}
    \label{fig:motivation}
    \vspace{-0.25in}
\end{figure}

Precise camera movement in cinematography is essential to set the scene for storytelling and to bootstrap multiple views in synthetic video generation. Digital video reshooting offers an accessible and powerful solution to digitally modify camera trajectories in post-production, as well as to improve model generalization in tasks, such as robotics, by expanding the range of training videos. 

However, developing robust video reshooting models for dynamic, non-rigid scenes is severely bottlenecked by the lack of large-scale, paired multi-view video data. To effectively train a video-to-video reshooting model, one inherently needs a pair of videos capturing the exact same action from different camera perspectives. To circumvent this scarcity, most existing approaches fall into two categories. First, some methods rely purely on large-scale synthetic datasets~\cite{bai2025recammaster, gcd}. While useful, the synthetic-to-real generalization gap is notoriously difficult to bridge. A model trained on rendered graphics struggles to generalize to the diverse domains of photorealistic video, animation, or complex dynamic interactions not seen in the training distribution, often producing significant visual artifacts (Fig.~\ref{fig:motivation}b).

Second, other methods attempt to build complex 4D representations from pretrained 3D/4D models~\cite{4d_recon_harley2025alltracker,4d_recon_hu2025depthcrafter,4d_recon_wang2024dust3r}. The simplest of these operate as \textit{anchor-only models}~\cite{ex4d}. As shown in Fig.~\ref{fig:motivation}(a), this approach is highly vulnerable to errors in the 3D/4D reconstruction, propagating artifacts in the anchor directly to the final output. Advanced techniques like \trajectorycrafter{}~\cite{yu2025trajectorycrafter} attempt to mitigate these artifacts by using the original source video as an additional input. However, these approaches remain constrained by the limited quality and high computational cost of their underlying 4D data pipelines.

In this paper, we propose a fundamentally different approach. Our core idea is a highly scalable, self-supervised strategy that generates a \textit{pseudo multi-view training triplet} (consisting of a source, anchor, and target video) from a single monocular video.

During training, we extract two distinct clips to serve as the source and target videos using smooth random walk trajectories that simulate dynamic camera motion, as shown in Fig.~\ref{fig:teaser}. We then synthetically generate the anchor video by forward-warping the first frame of the source video using a dense 2D tracker. This effectively simulates the distorted point-cloud-based inputs expected at inference. 

During inference, we convert the source video to a 4D point cloud, modify the perspective given a novel camera trajectory, and create an inference-time anchor video as geometric conditioning. The model must then predict a new target video given the anchor and the source videos.

Our training setup forces the base video model to learn a robust, 4D-aware understanding of scene dynamics. To reconstruct the target video, the model must learn to ignore reconstruction artifacts in the anchor and retrieve the correct texture from the source video. Because our cropping strategy introduces spatial misalignment and artificial occlusion, the model cannot simply copy the corresponding source frame. To fill dis-occluded regions, it is compelled to find the missing texture in a different temporal context within the source (Fig.~\ref{fig:teaser}). This spatial bottleneck acts as an information erasure mechanism, forcing the model to build an implicit 4D world prior by routing and re-projecting content across different times and pseudo-viewpoints.

Our contributions are three-fold:
\begin{itemize}[noitemsep, topsep=0pt, partopsep=0pt]
    \item First, we introduce a highly scalable, domain-agnostic self-supervised data pipeline that generates pseudo multi-view training triplets from monocular videos.
    \item Second, we propose a minimal and efficient architectural adaptation leveraging a pre-trained video diffusion model's self-attention modules. We pair this with a set of targeted training augmentations designed to ensure robust generalization to distorted, inference-time artifacts.
    \item Third, we demonstrate through extensive experiments that our approach achieves state-of-the-art temporal consistency and camera control across a diverse set of dynamic videos, significantly advancing the capabilities of video reshooting.
\end{itemize}
\section{Related Work}
\label{sec:related_work}

\subsection{Camera Controls for Video Generation Models}
The introduction of transformer-based diffusion models~\cite{peebles2023scalable,wan21,moviegen,seedance1,cogvideox} has catalyzed an explosion in video generation research and the development of methods that enable precise video control~\cite{vid_ctrl_fu20243dtrajmaster,vid_ctrl_guo2024sparsectrl,vid_ctrl_xing2024make,vid_ctrl_yin2023dragnuwa}. These additional control signals encompass user interactions like dragging~\cite{drag_geng2025motion,drag_jeong2024dreammotion,drag_jeong2024vmc,drag_zhao2024motiondirector}, explicit camera coordinates~\cite{camera_traj_he2024cameractrl,camera_traj_wang2024motionctrl,camera_traj_xiao2024trajectory,camera_traj_xu2024camco,camera_traj_yang2024direct,camera_traj_you2024nvs,camera_traj_zhang2025recapture}, and human pose estimation~\cite{human_animate_cheng2025wan,human_animate_gan2025humandit,human_animate_qin2023dancing,human_animate_wang2025unianimate}. Some techniques, particularly in video editing, use a source video directly as a conditional input~\cite{video_edit_cohen2024slicedit,video_edit_ouyang2024i2vedit,video_edit_wang2025videodirector}.

Within camera-controlled generation, many models utilize camera parameters~\cite{camera_traj_he2024cameractrl,gcd} or advocate for the integration of 3D priors, such as 3D meshes~\cite{bai2025recammaster,ex4d,yu2025trajectorycrafter} and point tracking~\cite{wang2025epic,jeong2025reangle}, to furnish the model with robust spatial cues. These techniques often involve converting the source image into a 3D representation to generate pseudo-anchors for guided text/image-to-video synthesis.

\begin{figure*}
    \centering
    \includegraphics[width=\linewidth]{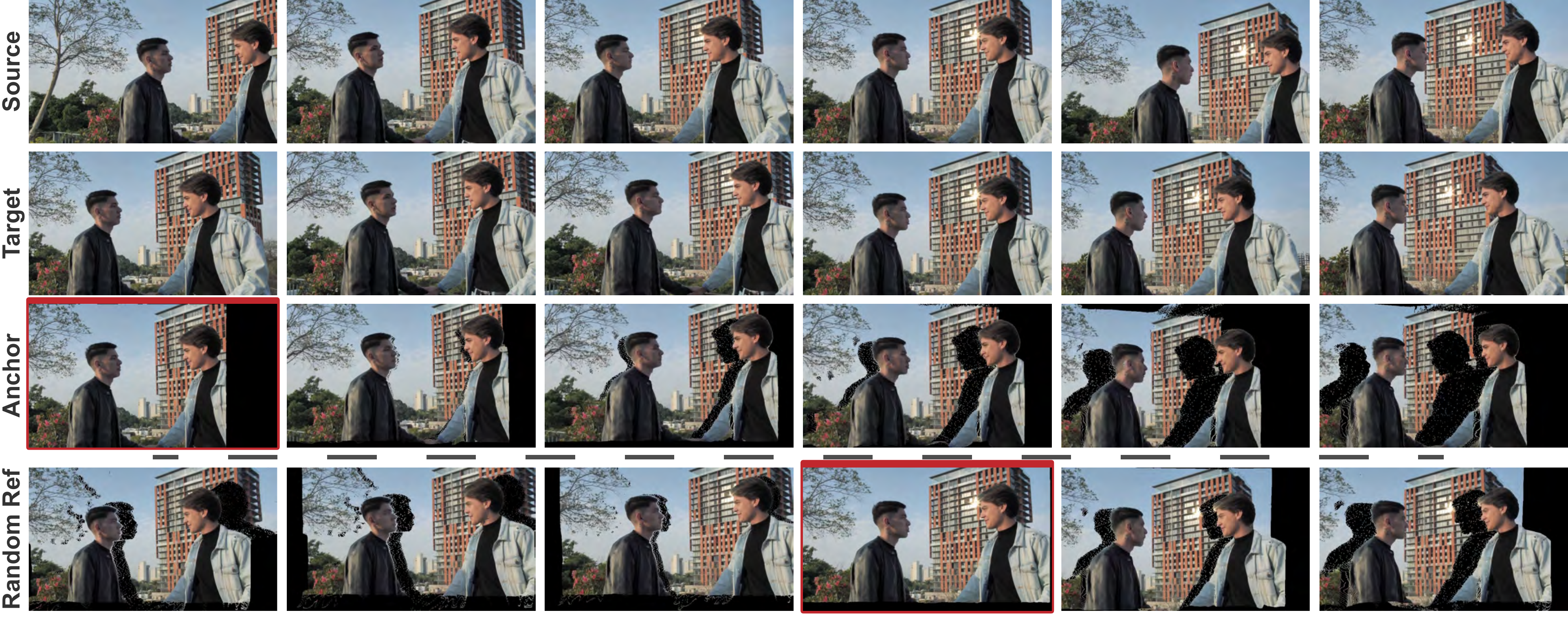}
    \vspace{-0.25in}
    \caption{
        \textbf{Our Pseudo Multi-View Triplet and Training Augmentations.}
        \textbf{Top three rows:} Our core training triplet consists of the \textit{Source Video} ($V_s$), the \textit{Target Video} ($V_t$), and the synthetically generated \textit{Anchor Video} ($V_a$). By default, $V_a$ is created by forward-warping a reference frame from $V_s$ to align with the $V_t$ trajectory (using a dense 2D tracker~\cite{4d_recon_harley2025alltracker}). This process incorporates augmentations like 3D-aware noise injection into the reference frame before warping, simulating inference-time artifacts.
        \textbf{Bottom row:} In practice, we randomly select the reference anchor frame for 2D tracking and warping. This reduces overfitting and improves anchor alignment for long sequences.
    }
    \label{fig:triplets}
    \vspace{-0.2in}
\end{figure*}

\subsection{Video-to-Video Generation and Re-shooting}
Camera-controlled video reshooting has been investigated in several recent works, including \recammaster{}~\cite{bai2025recammaster}, \exfd{}~\cite{ex4d}, and \trajectorycrafter{}~\cite{yu2025trajectorycrafter}. Methods relying purely on synthetic video data~\cite{bai2025recammaster, gcd} suffer from an inherent generalization gap. They typically fail on real-world samples or novel object interactions outside their training distribution. However, as we demonstrate, utilizing synthetic data sparingly alongside real-world data provides critical signals for extreme camera translations.

To address the synthetic-to-real gap, other methods employ 4D point tracking and depth estimation on in-the-wild videos~\cite{ex4d,jeong2025reangle,chen2025reconstruct}. Crucially, methods like EPiC~\cite{wang2025epic} rely solely on the projected anchor without querying the full source video. As the camera moves, the anchor accumulates artifacts, forcing the model to overfit to noisy textures and resulting in severe texture drift. Our method fundamentally differs by using the anchor strictly for geometric guidance while actively retrieving clean textures directly from the source video.

Furthermore, dynamic video reshooting must be clearly distinguished from static Novel View Synthesis (NVS). While concurrent self-supervised methods~\cite{mitchel2026true,jiang2025rayzer} excel at static NVS, they inherently fail to handle the non-rigid motion required for complex dynamic scenes. Additionally, while some approaches leverage 360-degree panoramic videos~\cite{xia2025panowan,luo2025beyond} to achieve angular diversity, high-quality dynamic 360-degree datasets remain relatively scarce. Our method instead unlocks the abundance of internet-scale standard perspective videos.

In contrast to existing constraints, our scalable hybrid methodology generates training pairs directly from monocular videos to teach implicit 4D spatiotemporal routing. By co-training this real-world data with a small fraction of multi-view synthetic data, our architectural adaptations facilitate strong alignment with extreme camera trajectories while maintaining high quality textures from source video.
\section{Methodology}
\label{sec:methodology}

In this work, we adapt a pre-trained video generation model for video reshooting. This task requires modifying the camera trajectory of an existing video while preserving its core content and visual style. We first define the overall task, then describe our novel self-supervised data pipeline, and finally detail the architectural adaptations and training augmentations required to implement our approach.

\begin{figure*}
    \centering
    \includegraphics[width=0.8\linewidth]{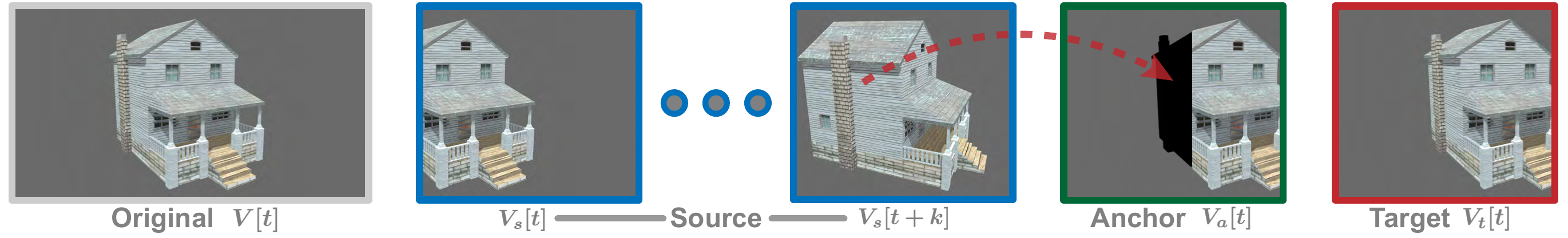}
    \vspace{-0.1in}
    \caption{
    \textbf{Implicit Spatiotemporal Reasoning in our Self-Supervised Setup.}
    Our training method forces the model to learn spatial structure from 2D data. To reconstruct the target $V_t$ at frame $t$, the model is given the disoccluded anchor $V_a[t]$. The corresponding source frame $V_s[t]$ may not contain the missing texture (e.g., the side of the building). The model is forced to find this texture in a different source frame, $V_s[t+k]$, where it is visible due to $V_s$'s independent camera motion. While this example shows a static 3D building for simplicity, performing this routing on dynamic videos forces the model to learn the scene's underlying complex 4D interactions.
    }
    \label{fig:implicit_3d}
    \vspace{-0.2in}
\end{figure*}

The video reshooting task requires synthesizing a high-fidelity target video ($V_{t}$) based on a new desired camera path. At inference time, our model is conditioned on two primary inputs: the source video ($V_{s}$), which serves as the high-quality texture and content reference, and the anchor video ($V_{a}$), which defines the target camera motion. This anchor video is typically pre-rendered from a 4D representation of the original scene.

Following recent approaches~\cite{yu2025trajectorycrafter,ex4d,jeong2025reangle,bai2024syncammaster,wang2025epic}, we utilize an anchor video as a primary conditioning signal. This choice is crucial for two main reasons:
\begin{description}[noitemsep, topsep=0pt, partopsep=0pt, leftmargin=0pt, style=unboxed, font=\normalfont\bfseries]
    \item [Intuitive Visual Guidance:] An anchor video offers a dense, intuitive visual representation of the target camera trajectory, which is empirically more effective for guiding generative models than explicit camera poses~\cite{wang2025epic}.
    \item [Self-Supervised Training Necessity:] It is essential for our novel self-supervised training strategy (Section~\ref{sec:self_supervision}). Our method generates pseudo-views from 2D crops without modifying underlying 3D camera parameters, making relative camera poses non-viable. The anchor provides the concrete spatiotemporal target needed for our pseudo multi-view pairs.
\end{description}

While the anchor video ($V_a$) is crucial for defining the target trajectory, it is inherently imperfect, suffering from artifacts like holes and inaccuracies caused by depth and pose estimation errors. The model's core task is to generate $V_{t}$ that faithfully follows the motion of $V_{a}$ while replacing its distorted content with clean, temporally consistent textures sourced from $V_{s}$. Training this two-stream model ideally requires large-scale triplets $(V_{s}, V_{a}, V_{t})$ with pristine $V_t$ ground-truth, which is logistically intractable. We address this with our novel self-supervised approach, generating these triplets from monocular videos alone.

\subsection{Self-Supervision from Monocular Video}
\label{sec:self_supervision}

To bypass the challenges of paired multi-view data acquisition, we synthesize the required $(V_s, V_a, V_t)$ training triplets solely from abundant monocular videos.

\subsubsection{Pseudo-View Triplet Generation}
\label{sec:pseudo_view_triplet}

From a single input video $V$, we employ a smooth video cropper to generate two distinct clips. These clips will form our source video ($V_s$) and target video ($V_t$). Each clip follows an independent smooth random-walk trajectory that emulates dynamic camera motion, as shown in Fig.~\ref{fig:teaser}. Each trajectory is generated by sampling an adaptive number of random control points within the frame boundaries, with the overall motion magnitude governed by a tunable scale parameter. A natural cubic spline is then fitted through these control points to produce a continuous crop trajectory.

We sample two trajectories using different random seeds, resulting in clips $(V_{s}, V_{t})$ that observe different regions of the same video. Because they are drawn from the same source, this effectively mimics time-synchronized distinct moving cameras. Importantly, the original video $V$ may already exhibit its own inherent camera motion. Our method naturally inherits this, capturing complex patterns such as orbits, pans, and zooms from in-the-wild footage.

Based on both clips, we synthetically generate the anchor video ($V_{a}$) to complete the training triplet. We generate $V_{a}$ by warping the first frame of the source video ($V_{s}[0]$) to align with the target video's trajectory ($V_{t}$). This warping process, illustrated in Fig.~\ref{fig:triplets}, is guided by an offset-aware flow field combining a dense tracking field (capturing 2D pixel motion) with a crop offset flow (capturing the change in pseudo-viewpoint). To obtain the dense tracking field, we use AllTracker~\cite{4d_recon_harley2025alltracker} on the original video $V$ before cropping, leveraging all intermediate frames for robust temporal correspondences. This combined flow is then used to forward-warp $V_{s}[0]$ over time using softmax splatting~\cite{softmax_splatting}:
\begin{equation}
V_{a}[t] = \text{SoftSplat}(V_s[0], F_{\text{comb}}(t)),
\label{eq:anchor_warping}
\end{equation}
where $V_a[t]$ is the $t$-th frame of the anchor video, $\text{SoftSplat}$ is the forward warping operator, and $F_{\text{comb}}(t)$ is our combined offset-aware flow field.  The use of forward warping is a deliberate choice, effectively simulating the artifacts, occlusions, and dis-occlusions characteristic of the point-cloud-based anchor videos used at inference.

\begin{figure*}
    \centering
    \includegraphics[width=0.8\linewidth]{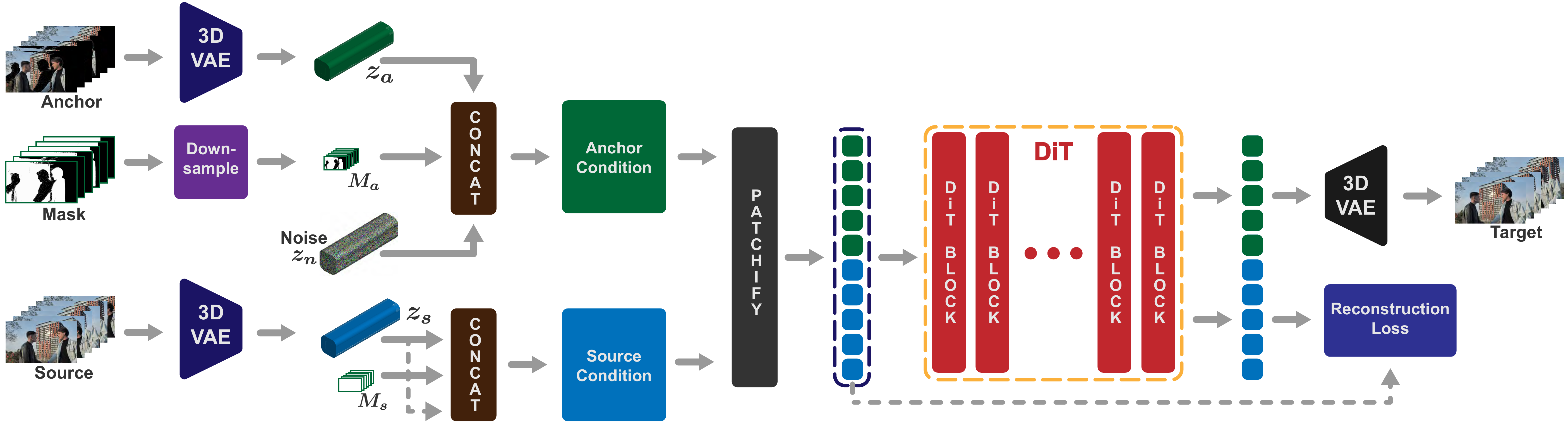}
    \vspace{-0.13in}
    \caption{\textbf{Overview of our Conditioning Architecture.} Our model adapts a pre-trained DiT-based I2V model~\cite{wan21}. \textbf{(1) VAE Encoding:} The Anchor video ($V_a$) and Source video ($V_s$) are independently encoded into latents by the VAE ($z_a, z_s$). \textbf{(2) Conditioning Setup:} The Anchor latent ($z_a$) is combined with a noise latent ($z_{n}$) and its corresponding mask ($M_a$). The Source latent ($z_s$) is duplicated (replacing $z_{n}$) and combined with an all-ones mask ($M_s$). \textbf{(3) DiT Processing:} The two conditioned streams are temporally concatenated and fed into the DiT blocks. \textbf{(4) Source Token Management \& Denoising:} After each denoising step, the output tokens corresponding to the Source are subjected to an auxiliary reconstruction loss (ensuring source content retention).}
    \label{fig:overview}
    \vspace{-0.2in}
\end{figure*}

\subsubsection{Implicit 4D-Aware Learning}
\label{sec:implicit_4d_learning}

This training setup directly forces the model to learn its inference-time task. It must reconstruct the high-quality, dynamic $V_{t}$ by mastering the following sub-tasks:

\begin{description}[noitemsep, topsep=0pt, partopsep=0pt, leftmargin=0pt, style=unboxed, font=\normalfont\bfseries]
    \item[Ignoring Anchor Artifacts] The model learns to treat $V_{a}$ strictly as a geometric guide, ignoring warping artifacts and relying on it only for structure.
    \item[Temporal Synchronization of Dynamics] By training on perfectly synchronized pseudo-views of real-world events, the model natively learns to map complex non-rigid motion frame-by-frame. This directly solves the synchronization failures commonly seen in models trained purely on synthetic data or static-dynamic split datasets.
    \item[Spatial Content Sourcing] The model is forced to find the correct, high-fidelity textures from $V_{s}$. Because $V_{s}$ and $V_{t}$ are not spatially aligned, the model must use its attention mechanism to route content from different spatial regions of the source.
    \item[Implicit 4D Reconstruction] As shown in Fig.~\ref{fig:implicit_3d}, generating a target frame often reveals hidden areas that are not visible in the corresponding source frame. To fill these missing regions, the model must search through the source video to find the required texture in a different frame. It then stitches this temporal information into the correct spatial position. While Fig.~\ref{fig:implicit_3d} illustrates this concept on a static building for simplicity, applying this "search and stitch" process to videos of moving subjects forces the model to track geometry across both space and time. This fundamentally teaches the model implicit 4D reconstruction.
    \item[Generative In-painting] For regions not visible in any $V_{s}$ frame, the model learns to leverage its generative priors to plausibly hallucinate the missing details.
\end{description}

This self-supervised strategy has key advantages over synthetic or 3D-annotated approaches. Our entire data pipeline relies only on a robust 2D dense tracker, making it domain-agnostic and applicable to virtually any monocular video. As a result, we can train on a vast corpus that includes photorealistic footage, animation, and generative art, exposing the model to diverse camera motions without restrictive domain limitations. Importantly, our training objective does not directly optimize for explicit 4D reconstruction; rather, dynamic 4D re-rendering emerges as a learned capability to solve this complex 2D-based routing task.

\subsection{Model Architecture and Conditioning}
\label{sec:architecture}

Having established our self-supervised training data (Section~\ref{sec:self_supervision}), we now detail the architectural adaptations required to finetune a pre-trained video diffusion model for our task, as illustrated in Figure~\ref{fig:overview}.

\subsubsection{Video Diffusion Models and DiT Conditioning}
\label{sec:arch_background}

Our approach adapts the WAN 2.2 I2V architecture~\cite{wan21}, a Diffusion Transformer (DiT). The base model is trained to predict the denoised latent from a noisy input, effectively reversing the diffusion process to generate high-quality frames. Conditional signals in DiT architectures can be integrated via channel concatenation, dedicated cross-attention layers, or as supplementary tokens processed by the full self-attention mechanism~\cite{bai2025recammaster, peebles2023scalable}. The chosen mechanism is critical for temporal modeling and Rotary Positional Embeddings (RoPE), which encode relative spatiotemporal positional information~\cite{bai2025recammaster, su2024roformer}. We leverage a token-based conditioning approach as detailed below.

\subsubsection{Source-Anchor Fusion with Offset RoPE}
\label{sec:offset_rope}

Our model leverages an anchor video ($V_a$) for geometric guidance and a source video ($V_s$) for high-fidelity texture. Instead of utilizing separate cross-attention layers for $V_s$, which often proves ineffective for this task~\cite{bai2025recammaster}, we process both $V_s$ and $V_a$ as tokens within the model's main self-attention mechanism.

Both $V_a$ and $V_s$ are independently encoded into latents ($z_a, z_s$) by the VAE. The Anchor condition combines $z_a$ with a noise latent ($z_n$) and its downsampled binary mask ($M_a$). The Source condition uses $z_s$ with an all-ones mask ($M_s$), ensuring all content is leveraged. These two prepared conditional streams are then temporally concatenated and fed into the DiT blocks.

Within each DiT block, 3D RoPE is applied. To enable robust flexible-length video generation, we implement an Offset RoPE. A large, fixed offset is applied only to the temporal positional embeddings of the $V_s$ tokens, effectively decoupling the source condition's perceived position from the target's temporal position. While output source tokens do not directly form $V_t$, an auxiliary reconstruction loss is applied during training between the output source tokens and their original clean latents ($z_s$). This encourages the source pathway to preserve high-fidelity content from $V_s$.

\begin{figure}
    \centering
    \includegraphics[width=\linewidth]{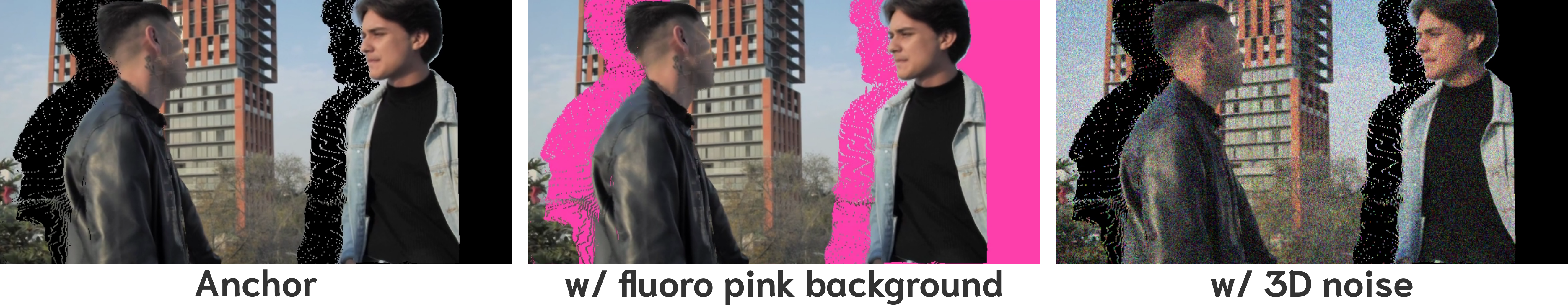}
    \vspace{-0.25in}
    \caption{This figure demonstrates key augmentations applied during Anchor Video ($V_a$) generation, shown on a single frame. From left to right: the default anchor with a black masked background; an anchor using a fluorescent pink background for masked regions; and an anchor with 3D-aware noise coherently injected into its reference frame. These techniques improve model robustness and help mitigate artifacts in challenging scenes (Section~\ref{sec:technical_augmentations}).}
    \label{fig:ablations_input}
    \vspace{-0.2in}
\end{figure}

\begin{figure*}
    \centering
    \includegraphics[width=\linewidth]{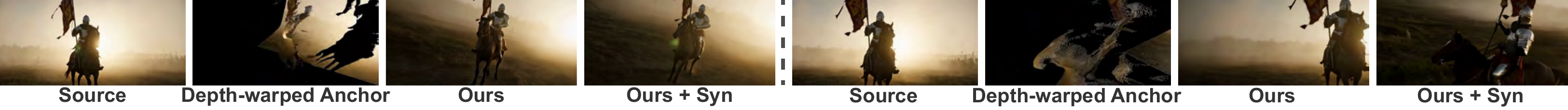}
    \vspace{-0.27in}
    \caption{
        We evaluate models trained with (Ours + Syn) and without (Ours) the 15\% synthetic data mixture. \textbf{Left:} Under moderate camera motion, both models successfully follow the anchor video, demonstrating that our self-supervised monocular pipeline natively learns robust 4D structures. \textbf{Right:} For extreme out-of-distribution camera rotations, the monocular-only model struggles to align with the anchor. Conversely, the hybrid model accurately follows the extreme trajectory while preserving high-fidelity textures, confirming the synthetic mixture provides important priors for extreme 6DoF camera generalization.
    }
    \label{fig:hybrid_ablation}
    \vspace{-0.25in}
\end{figure*}

\subsubsection{Technical Augmentations for Robust Training}
\label{sec:technical_augmentations}

Beyond our core architecture, we integrate several technical augmentations to enhance robustness against inference-time artifacts (Fig.~\ref{fig:ablations_input}). Detailed configurations are provided in the Supplementary Material.

\begin{description}[noitemsep, topsep=0pt, partopsep=0pt, leftmargin=0em, style=unboxed, font=\normalfont\bfseries]
    \item[Source Token Reconstruction Loss] An auxiliary loss applied to ensure content retention from the source video.
    \item[Fluorescent Anchor Background] We use a high-contrast background, such as fluorescent pink, in masked-out regions of the anchor video to provide a clear boundary signal. This is specially effective for dark scenes.
    \item[Random Anchor Reference Frame Warping] Randomly selecting the reference frame for anchor warping prevents model bias towards early frames.
    \item[3D-Aware Noise] Gaussian noise is applied directly to the anchor's reference frame prior to warping. This noise moves coherently with the underlying structure, suppressing simple texture copying from anchor while strictly preserving geometric guidance.
\end{description}

\section{Implementation Details}
\label{sec:implementation}

\subsection{Dataset Creation and Hybrid Strategy}
\label{sec:dataset_creation}

Our primary training data consists of approximately 100,000 clips extracted from 30,000 high-resolution, in-the-wild monocular videos using our smooth random-walk cropping pipeline (Sec.~\ref{sec:pseudo_view_triplet}). While training exclusively on this data yields high-quality outputs for standard trajectories, we observed limited generalization to extreme camera motions, such as 120-degree orbital camera rotations.

Because manually curating real-world videos with sweeping camera motion is difficult, we introduce a hybrid strategy. We augment our training pool with a 15\% mixture of paired multi-view synthetic data from a random subset of \recammaster{}~\cite{bai2025recammaster}. As analyzed in our ablation studies (Sec.~\ref{sec:ablations}), this deliberate combination allows the model to learn extreme camera path alignment from the synthetic data while securely grounding its complex physics and textures in the real-world dataset. We highlight this in Fig.~\ref{fig:hybrid_ablation}.

The processing for this synthetic mixture mirrors our monocular pipeline, utilizing forward warping to generate anchor videos by tracking relative motion between the two video sequences. Complete processing details and visualizations are deferred to the Supplementary Material.

\subsection{Architectural Configurations}
\label{sec:arch_configs}

We build our architecture upon the \textit{Wan2.2-14B Mixture-of-Experts (MoE)} video diffusion model~\cite{wan21}. To process dual video inputs via custom token-based conditioning, both the anchor ($V_a$) and source video ($V_s$) are encoded by the pre-trained causal 3D VAE into $T_L = 20$ latent frames.

The Anchor conditioning stream concatenates the VAE-encoded anchor latent ($z_a$) with a noise latent ($z_n$) and a downsampled binary mask ($M_a$) indicating valid generation regions. The Source conditioning stream duplicates the source latent ($z_s$) to replace the noise channels, concatenating it with an all-ones mask ($M_s$) to leverage all source content. These prepared streams are temporally concatenated, yielding a token sequence twice as long as standard inference ($2 \times T_L$ latent frames).

To distinguish the positional context of these streams, we apply an Offset RoPE. A constant offset of 50 is added to the 3D-RoPE temporal embeddings of the source tokens. Because this offset significantly exceeds our maximum training latent frames ($T_L = 20$), it strictly decouples the source condition's perceived temporal position from the target's active denoising trajectory.

\subsection{Training Details}
\label{sec:training_details}

The high-noise and low-noise MoE experts are trained independently. We empirically observe that the high-noise model exerts significantly greater influence on camera motion due to its operation during early denoising timesteps. Consequently, all anchor-video ablations are conducted exclusively on the high-noise model. The low-noise model is trained using standard black-background anchors without the source reconstruction loss, and is shared across all ablations to ensure fair comparisons.

To prevent the model from directly copying anchor video textures, we inject 3D-aware Gaussian noise into the reference frame's RGB values prior to forward warping. The noise magnitude is sampled from a uniform distribution between [0, 0.5] per channel on normalized images.

All variants are trained for 2K steps with a batch size of 24, a 1e-5 learning rate, and the AdamW optimizer ($\beta_1=0.9$, $\beta_2=0.999$). To efficiently adapt the 14B parameter backbone, we apply a rank-512 LoRA on the attention and feed-forward layers and fully train the patchify layer. When incorporating the source reconstruction loss for high-fidelity texture routing, we compute an L1 loss between the output source tokens and the clean source latent. We backpropagate on the total objective $\mathcal{L}_{\text{total}} = \mathcal{L}_{\text{MSE}} + \alpha \,\mathcal{L}_{\text{reference}}$, with $\alpha$ set to 0.1.

\section{Experiments}
\label{sec:experiments}

\begin{table*}[t]
\label{table-main-comparison}
	\begin{center}
\vspace{-0.30cm}
		\caption{Quantitative comparison with SOTA methods on VBench, temporal consistency, camera accuracy, and view synchronization.}
\vspace{-0.30cm}
		\label{tab_merged_eval}
		\setlength\tabcolsep{3.2pt} % Keep small for fitting
 	\centering
 	\resizebox{\textwidth}{!}{ % Use \textwidth to fit the wide table
		\begin{tabular}{lcccccc|c|cc|ccc}
			\toprule
			\multirow{2}{*}{Method}
            & \multicolumn{6}{c|}{VBench Quality}
            & \multicolumn{1}{c|}{Temporal Consistency}
            & \multicolumn{2}{c|}{Camera Accuracy}
            & \multicolumn{3}{c}{View Synchronization} \\
     	\cmidrule(lr){2-7}
     	\cmidrule(lr){8-8}
     	\cmidrule(lr){9-10}
     	\cmidrule(lr){11-13}
            & \makecell[c]{Aesthetic $\uparrow$} & \makecell[c]{Imaging $\uparrow$} & \makecell[c]{Flickering $\uparrow$} & \makecell[c]{Smoothness $\uparrow$} & \makecell[c]{Subject $\uparrow$} & \makecell[c]{Background $\uparrow$}
			& CLIP-F $\uparrow$
			& RotErr $\downarrow$
			& TransErr $\downarrow$
			& Mat. Pix $\uparrow$
			& FVD-V $\downarrow$
			& CLIP-V $\uparrow$ \\
		\midrule

	\trajectorycrafter{} (49 frames) & 52.69 & \textbf{59.67} & 96.97 & 99.03 & 93.78 & 95.13 & 98.80 &  \textbf{2.26} & 3.03 & 1851.80 & 582.56 & 92.40 \\
	\textbf{Ours} (49 frames) & \textbf{52.72} & 57.81 & \textbf{97.43} & \textbf{99.24} & \textbf{95.09} & \textbf{95.62} & \textbf{99.01} & 2.61 & \textbf{2.73} & \textbf{2737.65} & \textbf{488.22} & \textbf{94.96} \\
    \midrule
    \midrule
	\recammaster{} & 48.71 & 52.61 & \textbf{97.57} & \textbf{99.26} & 88.57 & 90.65 & 98.49 & 11.29 & 19.59 & 1314.00 & 732.52 & 88.91 \\
	\exfd{} & 49.72 & 55.76 & 97.46 & 99.08 & 91.51 & 94.78 & 98.94 & 3.94 & \textbf{4.21} & 2188.98 & 685.63 & 89.77 \\
	% \textbf{Ours} & 53.05 & 58.39 & 97.42 & 99.20 & 93.43 & 95.11 & 99.04 & 3.27 & 4.92 & 2636.07 & 595.39 & 92.94 \\
    % \midrule
    % \textbf{Ours w/o Source Video} & \textbf{53.97} & \textbf{58.92} & 97.13 & 98.99 & 93.12 & 95.03 & 98.95 & 4.82 & 11.65 & 226.01 & 662.09 & 89.97 \\
    \textbf{Ours} & \textbf{52.85} & \textbf{58.64} & 97.37 & 99.21 & \textbf{93.43} & \textbf{95.24} & \textbf{99.03} & \textbf{2.76} & 4.23 & \textbf{2720.83} & \textbf{586.24} & \textbf{93.16} \\
\bottomrule
		\end{tabular}
 	}
	\end{center}

    \vspace{-0.15in}
\end{table*}

\begin{figure*}
    \centering
    \includegraphics[width=\linewidth]{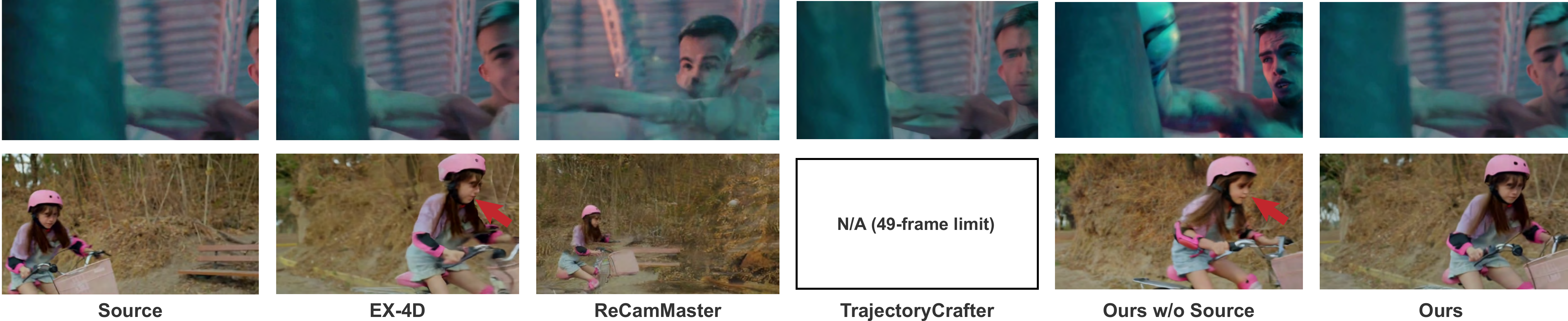}
    \vspace{-0.2in}
    \caption{We compare against other state-of-the-art approaches on the test set~\cite{lin2024open}. Existing approaches struggle to handle complex camera motions (e.g., shaky camera in the first example) and intricate details such as facial features in the second example (see supp. video).
    }
    \label{fig:comparisons}
    \vspace{-0.15in}
\end{figure*}

\subsection{Evaluation Setup} % Renamed for broader scope as it will include metrics, baselines etc.
\label{sec:eval_setup}

\textbf{Dataset.}
For evaluation, we constructed a representative set of 100 five-second videos from the publicly available Opensora-mixkit dataset~\cite{lin2024open}, each at 16 fps and 480p resolution. This evaluation set was systematically sampled to ensure a balanced representation of semantic concept coverage, camera motion diversity, and a mix of occluded and dis-occluded regions. More details are provided in the Supplementary Material.

\noindent\textbf{Metrics.} 
We evaluate performance across three dimensions. \textit{Camera accuracy} is measured via Rotational (RotErr) and Translational Error (TransErr) from extracted poses~\cite{camera_traj_he2024cameractrl, huang2025vipe}. \textit{View synchronization} is quantified using GIM average matching pixels (Mat. Pix.)~\cite{shen2024gim}, Fréchet Video Distance against the source video (FVD-V)~\cite{eval_unterthiner2019fvd}, and source-target frame similarity (CLIP-V). Finally, \textit{overall video quality} and temporal consistency are assessed using the VBench suite~\cite{eval_huang2024vbench} and adjacent-frame similarity (CLIP-F). More details in the Supplementary Material.

\subsubsection{Baseline Methods}
\label{sec:baselines}

We compare our approach against recent state-of-the-art video reshooting methods. We evaluate against \recammaster{}~\cite{bai2025recammaster}, which is conditioned on a source video and a target camera trajectory. We also compare against \exfd{}~\cite{ex4d}, an approach that solely relies on anchor videos, and \trajectorycrafter{}~\cite{yu2025trajectorycrafter}, which is conditioned on both anchor videos and a source video. For a fair comparison with \trajectorycrafter{}, which natively generates 49-frame videos, all metrics are computed on the first 49 frames of both their outputs and our corresponding generations.

\subsection{Comparisons with State-of-the-Art Methods}
\label{sec:comparisons}

Our approach consistently outperforms existing state-of-the-art methods in dynamic video reshooting. Quantitatively, as detailed in Table~\ref{tab_merged_eval}, our model demonstrates leading performance across most metrics. We achieve superior overall video quality, temporal consistency, and view synchronization while maintaining highly competitive camera accuracy and robust geometric fidelity against all baselines.

Qualitatively, Figure~\ref{fig:comparisons} illustrates two dynamic scenes from the test set to compare our output against existing baselines. The output quality of \recammaster{} is notably poor, which is partially due to the publicly released checkpoint being a limited 1.3B parameter model. For anchor-only methods like \exfd{}, the lack of direct source video conditioning causes the warping artifacts present in the anchor video to carry over directly into the generated output. Similarly, \trajectorycrafter{} struggles to maintain high visual fidelity due to limitations within its underlying training setup. To further validate the necessity of our architecture, we visualize our model trained without source video conditioning (`Ours w/o Source'). Much like \exfd{}, omitting the source video prevents the model from capturing and preserving the intricate details of the original footage.

Consequently, existing methods struggle with complex camera motions (e.g., shaky footage) and fine facial details. In contrast, our full method leverages implicit 4D routing to handle these challenges, producing high-fidelity outputs that preserve the original content. More comparisons on challenging trajectories are in the Supplementary Material.

\begin{figure}[ht]
    \centering
    \includegraphics[width=\linewidth]{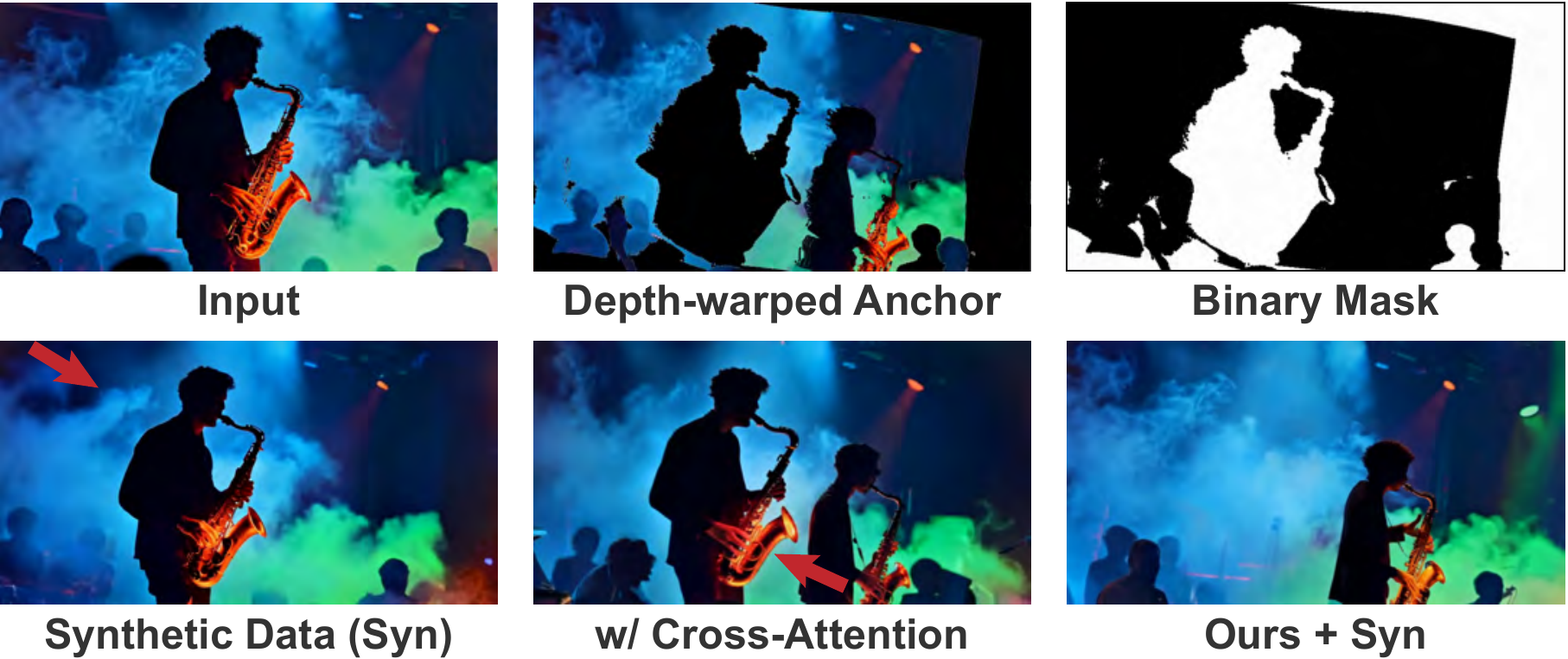}
    \vspace{-0.15in}
    \caption{We evaluate critical architectural and data components on a complex scene featuring moving smoke and colored lighting. The \textit{Synthetic Data Only} model fails to capture the intricate dynamics of the smoke. The \textit{Cross-Attention} model fails to preserve fine source details, such as the texture of the saxophone. Both baselines struggle to accurately align with the anchor condition, resulting in structural hallucinations. In contrast, our model faithfully tracks the anchor while preserving high-fidelity textures and complex scene dynamics.}
    \label{fig:ablation_comp}
    % \vspace{-0.15in}
\end{figure}
\begin{table}[ht]
\begin{center}
     \vspace{-0.10cm}
     \caption{Quantitative ablation of our training strategies, evaluating camera accuracy and view synchronization.}
     \vspace{-0.20cm}
     \label{tab_quality_eval}
     \setlength\tabcolsep{3.2pt}
     \centering
     \resizebox{\linewidth}{!}{
     \begin{tabular}{l|cc|ccc}
     \toprule
     \multirow{2}{*}{Method}
 & \multicolumn{2}{c|}{Camera Accuracy}
 & \multicolumn{3}{c}{View Synchronization} \\
 \cmidrule(lr){2-3}
 \cmidrule(lr){4-6}
 & RotErr $\downarrow$
 & TransErr $\downarrow$
 & Mat. Pix $\uparrow$
 & FVD-V $\downarrow$
 & CLIP-V $\uparrow$ \\
 \midrule

 Baseline (w/ self-attn)
 & 3.27 & 4.92 & 2636 & 595.39 & 92.94 \\
 \midrule
 \quad - Source Video
 & 4.82 & 11.65 & 226 & 662.09 & 89.97 \\
 \midrule

 \quad + Gaussian Noise in Latent
 & 2.81 & 5.05 & 2586 & 605.93 & 91.70 \\
 
 \quad + 3D Noise in Anchor
 & 2.49 & 4.95 & 2624 & 598.67 & 92.88 \\
 \midrule

\quad w/ cross-attn
        & 3.53 & 4.31 & 1766 & 626.37 & 91.64 \\
        
\quad + Auxiliary Loss
        & 2.76 & 4.98 & 2627 & 562.94 & 92.93 \\

\quad + LoRA
        & 2.85 & 4.17 & 2615 & 578.91 & 92.98 \\

\quad + Random Query
        & 3.36 & 4.52 & 2618 & 571.74 & 92.83 \\
        
\quad + Fluorescent Background Anchor
 & 3.16 & 4.78 & 2627 & 571.06 & 92.68 \\

 \midrule
 w/ Synthetic Data (Syn) & 3.70 & 5.04 & 1746 & 608.03 & 91.86 \\
 w/ Monocular Videos (\textbf{Ours}) & 2.76 & 4.23 & 2720 & 586.24 & 93.16 \\
 \textbf{Ours + Syn} & 3.36 & 3.66 & 2577 & 587.91 & 92.64 \\
 \bottomrule
 \end{tabular}
 }
\end{center}
    \vspace{-0.25in}
\end{table}

\subsection{Ablations}
\label{sec:ablations}

We conduct targeted ablation studies to quantify the impact of our core architectural choices and training data strategies against a defined baseline. Our baseline configuration uses a black background for the anchor video, processes the source video via token concatenation through the self-attention layers, and is trained exclusively on monocular videos without any additional augmentations. Detailed quantitative results are presented in Table~\ref{tab_quality_eval}, alongside some qualitative comparisons in Figure~\ref{fig:ablation_comp}.

\noindent\textbf{Hybrid Dataset.} We validate our data strategy by evaluating models trained exclusively on synthetic data (`w/ Synthetic Data (Syn)'), purely on monocular videos (`Ours'), and our 15\% mixture (`Ours + Syn'). While the synthetic-only model achieves reasonable camera alignment, it severely degrades texture fidelity. Qualitatively (Figure~\ref{fig:ablation_comp}), the synthetic-only baseline fails to capture complex real-world dynamics, such as smoke moving under colored lighting, and struggles with anchor alignment in complex scenes, leading to structural hallucinations. In contrast, our full hybrid approach (`Ours + Syn') seamlessly preserves these intricate dynamics, confirming that grounding the model in real-world triplets is essential for physical realism. Furthermore, as demonstrated in Figure~\ref{fig:hybrid_ablation}, while the purely monocular model (`Ours') excels under moderate motion, incorporating the synthetic mixture (`Ours + Syn') is critical for enabling robust geometric control during extreme, out-of-distribution camera trajectories.

\noindent\textbf{Conditioning Architecture.} We evaluate the integration of the source video condition by comparing our token concatenation strategy against a dedicated cross-attention mechanism (`w/ cross-attn'). Cross-attention performs significantly worse across all quantitative metrics. Visually (Figure~\ref{fig:ablation_comp}), this variant fails to reliably pass fine details from the source video to the generated output, completely losing the correct texture on the saxophone. It also exhibits poor anchor alignment. In contrast, our token concatenation approach natively leverages the pre-trained self-attention layers, yielding vastly superior view synchronization and structural accuracy.

\noindent\textbf{Technical Augmentations.} Beyond core architecture, we validate our technical augmentations in Table~\ref{tab_quality_eval}. Key findings confirm the critical roles of our fluorescent anchor background, 3D-aware noise injection for strict geometric guidance without texture copying, the auxiliary source token reconstruction loss for content fidelity, LoRA fine-tuning for generalization, and random anchor reference frame warping for robustness.

\section{Conclusion and Future Work}
\label{sec:conclusion}

We introduced a novel self-supervised framework for in-the-wild video reshooting that addresses the severe scarcity of paired multi-view data. By generating pseudo multi-view training triplets solely from monocular videos, our approach bypasses the need for explicit 3D annotations. We achieve this through a token-concatenation architecture within a DiT-based model, utilizing Offset RoPE and targeted augmentations to effectively route dynamic source textures. Extensive evaluations demonstrate that our method achieves state-of-the-art temporal consistency and geometric control, confirming that the model natively learns implicit 4D structures directly from 2D data. 

While our approach significantly advances dynamic reshooting, we identify key areas for future exploration. First, concatenating source tokens doubles the sequence length, impacting generation speed. Because source video information remains static across diffusion timesteps, future work could explore sophisticated KV-caching mechanisms to largely eliminate this computational overhead. Second, extreme trajectories that move entirely outside the original scene's boundaries yield blank anchor videos, removing geometric guidance. To resolve this, future iterations could employ a hybrid conditioning scheme leveraging our synthetic dataset mixture to co-train the model on both visual anchors and explicit camera poses. Finally, to further mitigate blank anchors and even more intricate camera paths, an autoregressive framework could be explored where the anchor condition is generated by forward-warping the previously generated target frame. This would naturally improve conditioning stability, representing a critical step toward true 4D generative world models.

% % WARNING: do not forget to delete the supplementary pages from your submission 
% % \input{sec_workshop/X_suppl}
% {
%     \small
%     \bibliographystyle{ieeenat_fullname}
%     \bibliography{main}
% }

% % % WARNING: do not forget to delete the supplementary pages from your submission 
% \input{sec_workshop/X_suppl}

{
    \small
    \bibliographystyle{ieeenat_fullname}
    \bibliography{main}
}

\clearpage
% \setcounter{page}{1}

% \begin{document}

\twocolumn[{
\renewcommand\twocolumn[1][]{#1}

\maketitlesupplementary
\begin{center}
  \centering
  % \vspace{-0.2in}
    \includegraphics[width=\linewidth]{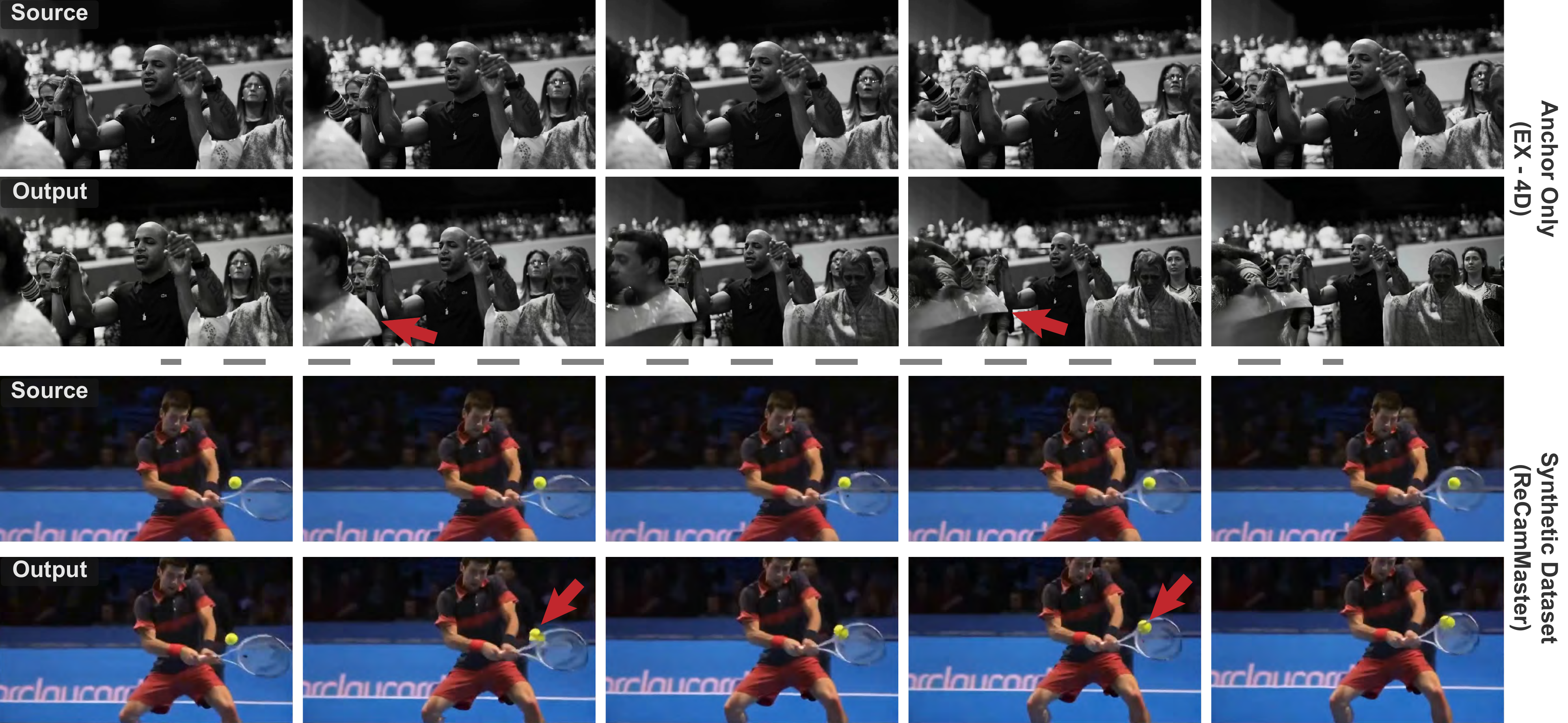}
    % \vspace{-0.25in}
\captionof{figure}{\textbf{Motivation: Illustrating Failure Modes in Video Reshooting.} This figure highlights common limitations of existing video reshooting methods, motivating our approach.
\textbf{(Rows 1-2) Anchor-Only Artifacts:} Given a high-quality source video (Row 1), an anchor-only method like \exfd{}~\cite{ex4d} (shown in Row 2) often generates significant ghosting and content inconsistencies. This occurs when the model relies solely on an imperfect anchor video ($V_a$) for geometric guidance and struggles to hallucinate complex details not fully represented in $V_a$.
\textbf{(Rows 3-4) Loss of Detail in Synthetic-Data Models:} For a given source video (Row 3), synthetic-data trained models, such as \recammaster{}~\cite{bai2025recammaster} (Row 4), can fail to capture intricate spatio-temporal details and object dynamics. In this example, \recammaster{} incorrectly deforms the tennis ball as it moves across the racket.
\textit{Note on \recammaster{}:} While \recammaster{} demonstrates impressive capabilities in general camera-controlled video generation, this result (obtained from the demo videos on their official project website) highlights that challenges with intricate object dynamics persist even in their best demonstrations, and similar artifacts are evident in their other results.}
    \label{fig:supp_motivation}
\end{center}
}
]

\begin{figure*}
    \centering
    \includegraphics[width=\linewidth]{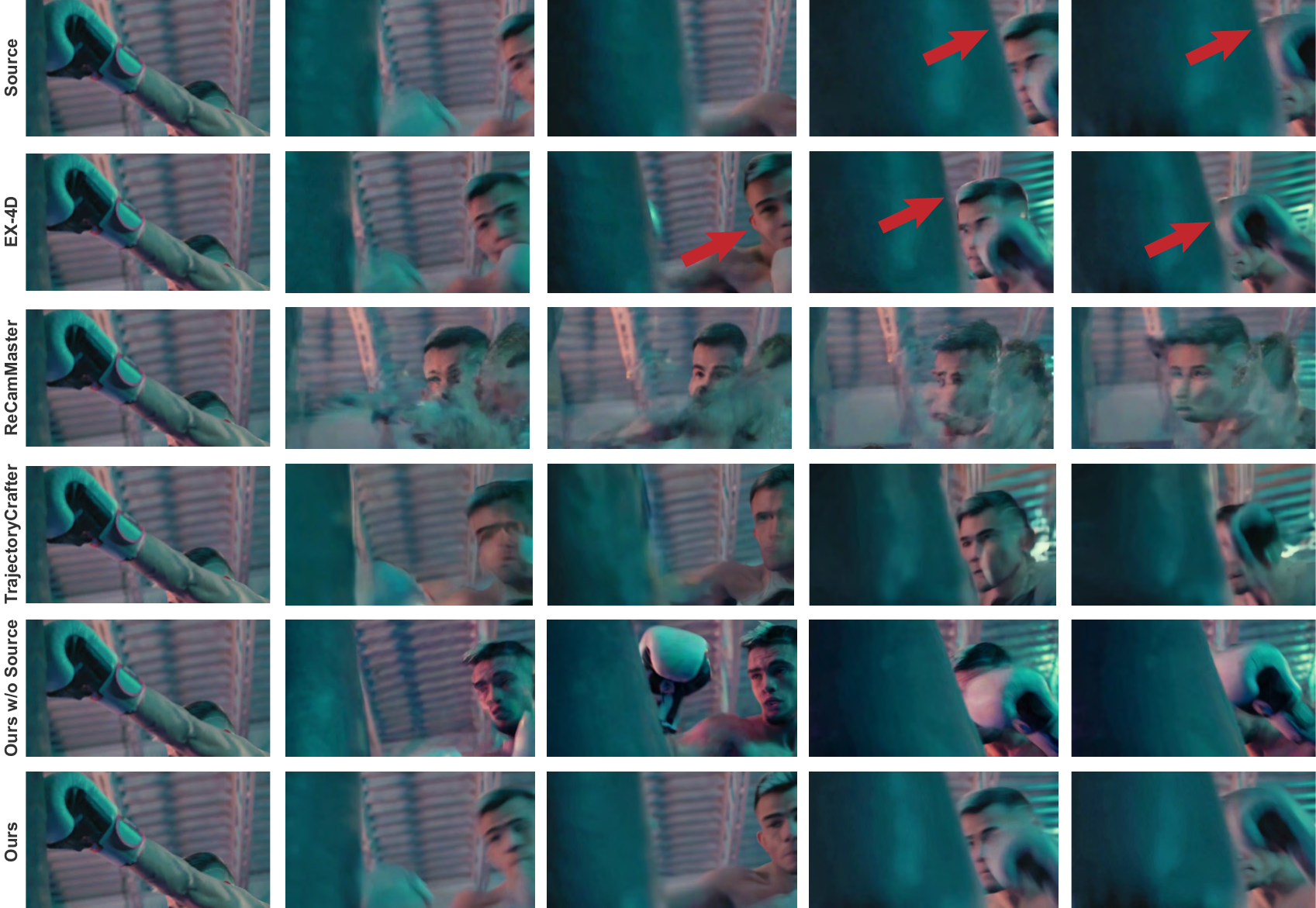}
    \vspace{-0.25in}
\caption{\textbf{Extended Qualitative Comparisons.} Each row displays sample frames from generated videos. Arrows indicate characteristic artifacts in baseline methods, such as loss of fine detail, blurring, or texture distortion. In contrast, our approach consistently demonstrates superior fidelity, accurately reproducing small details and effectively preserving intricate textures from the source video.}
    \label{fig:supp_comparison1}
    \vspace{-0.2in}
\end{figure*}

\begin{figure*}
    \centering
    \includegraphics[width=\linewidth]{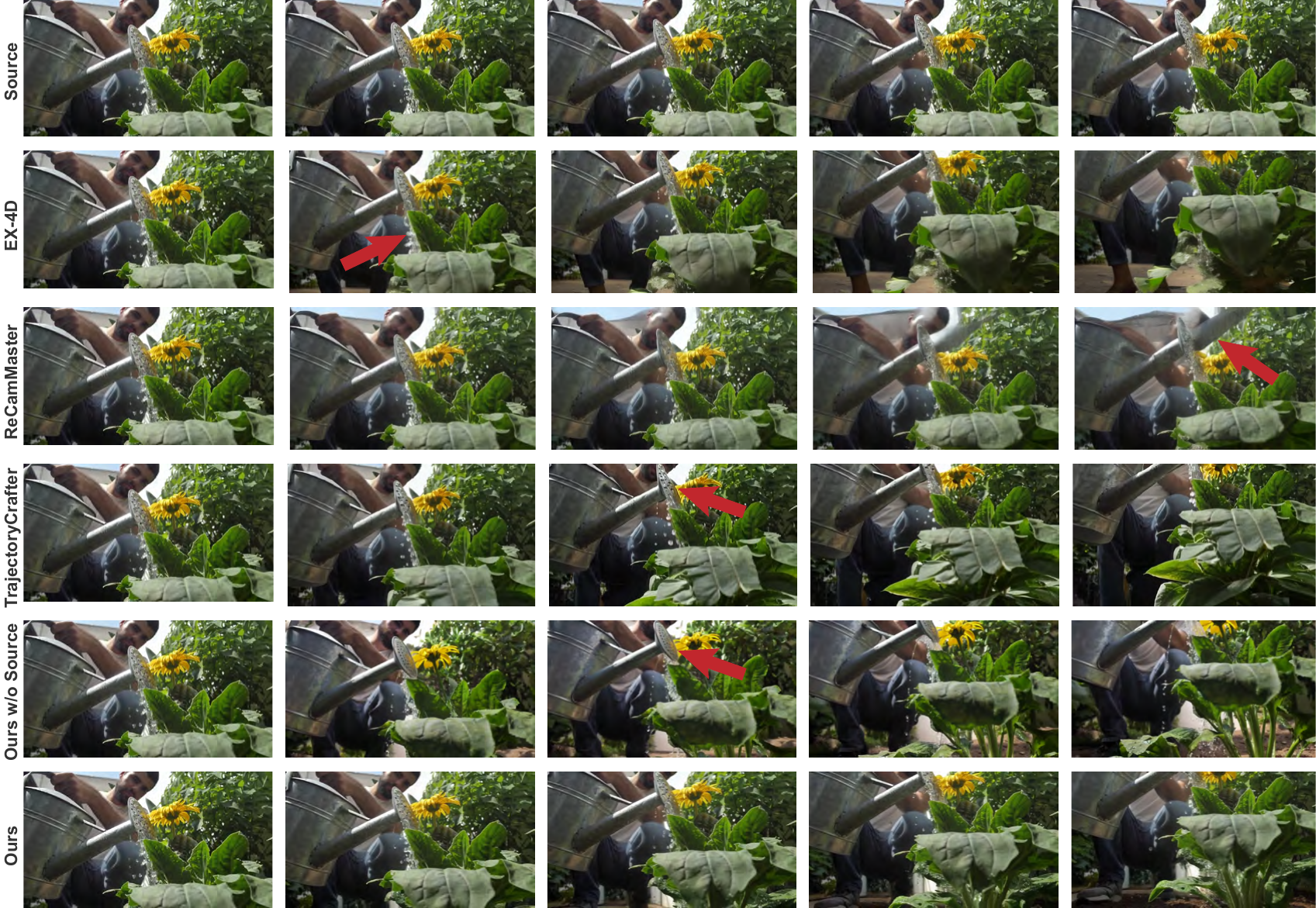}
    \vspace{-0.25in}
\caption{\textbf{Extended Qualitative Comparisons.} Each row displays sample frames from generated videos. Arrows indicate characteristic artifacts in baseline methods, as noted previously. Our approach maintains robust geometric fidelity and superior perceptual quality.}
    \label{fig:supp_comparison2}
    \vspace{-0.2in}
\end{figure*}

\begin{figure*}
    \centering
    \includegraphics[width=\linewidth]{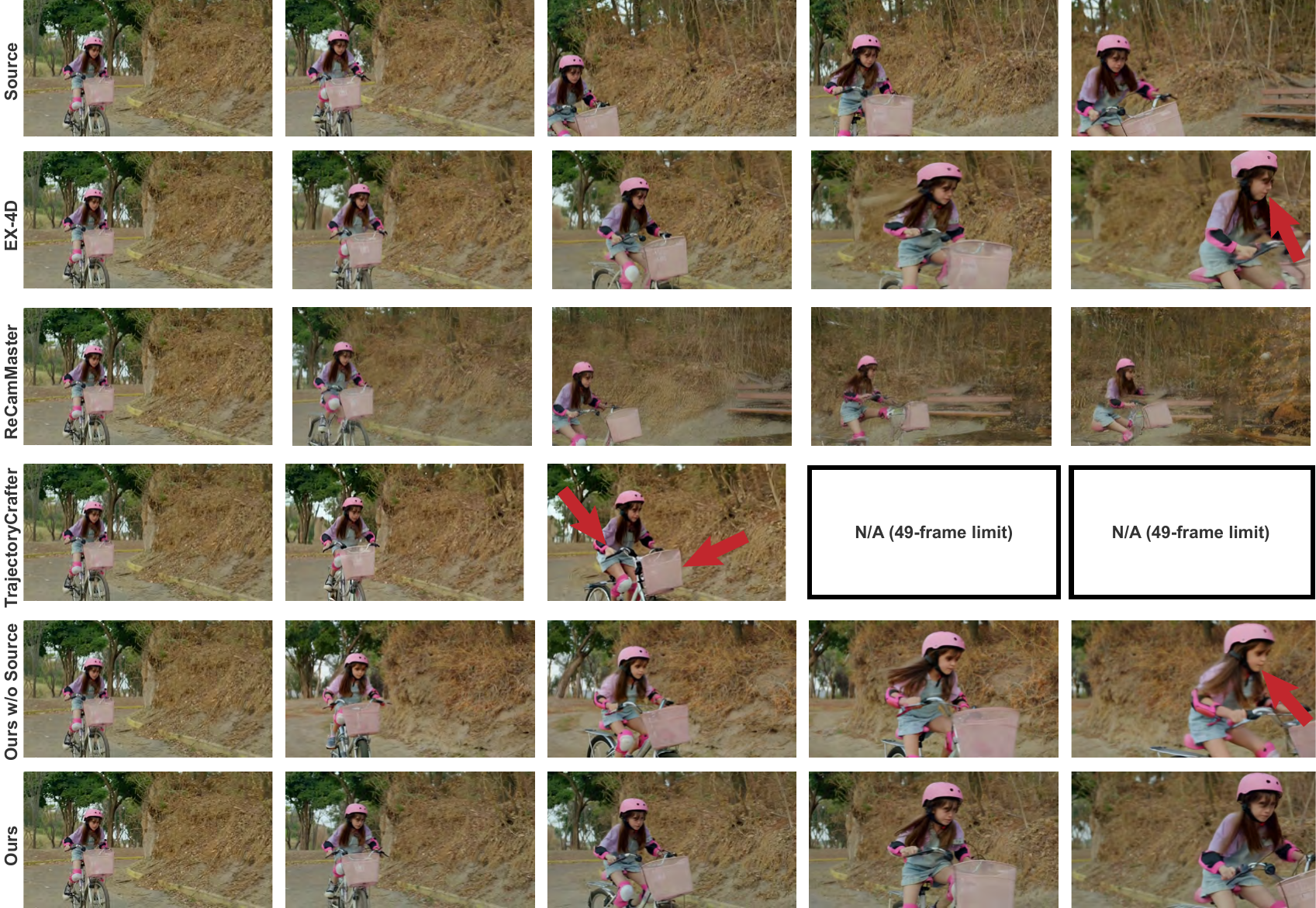}
    \vspace{-0.25in}
\caption{\textbf{Extended Qualitative Comparisons.} Additional examples demonstrating our model's ability to consistently reproduce fine details and preserve intricate textures from the source video across diverse scenes.}
    \label{fig:supp_comparison3}
    \vspace{-0.2in}
\end{figure*}

% \begin{figure*}
%     \centering
%     \includegraphics[width=\linewidth]{figures/supp_comp4.pdf}
%     \vspace{-0.25in}
% \caption{}
%     \label{fig:supp_comparison2}
%     \vspace{-0.2in}
% \end{figure*}

\begin{figure*}
    \centering
    \includegraphics[width=\linewidth]{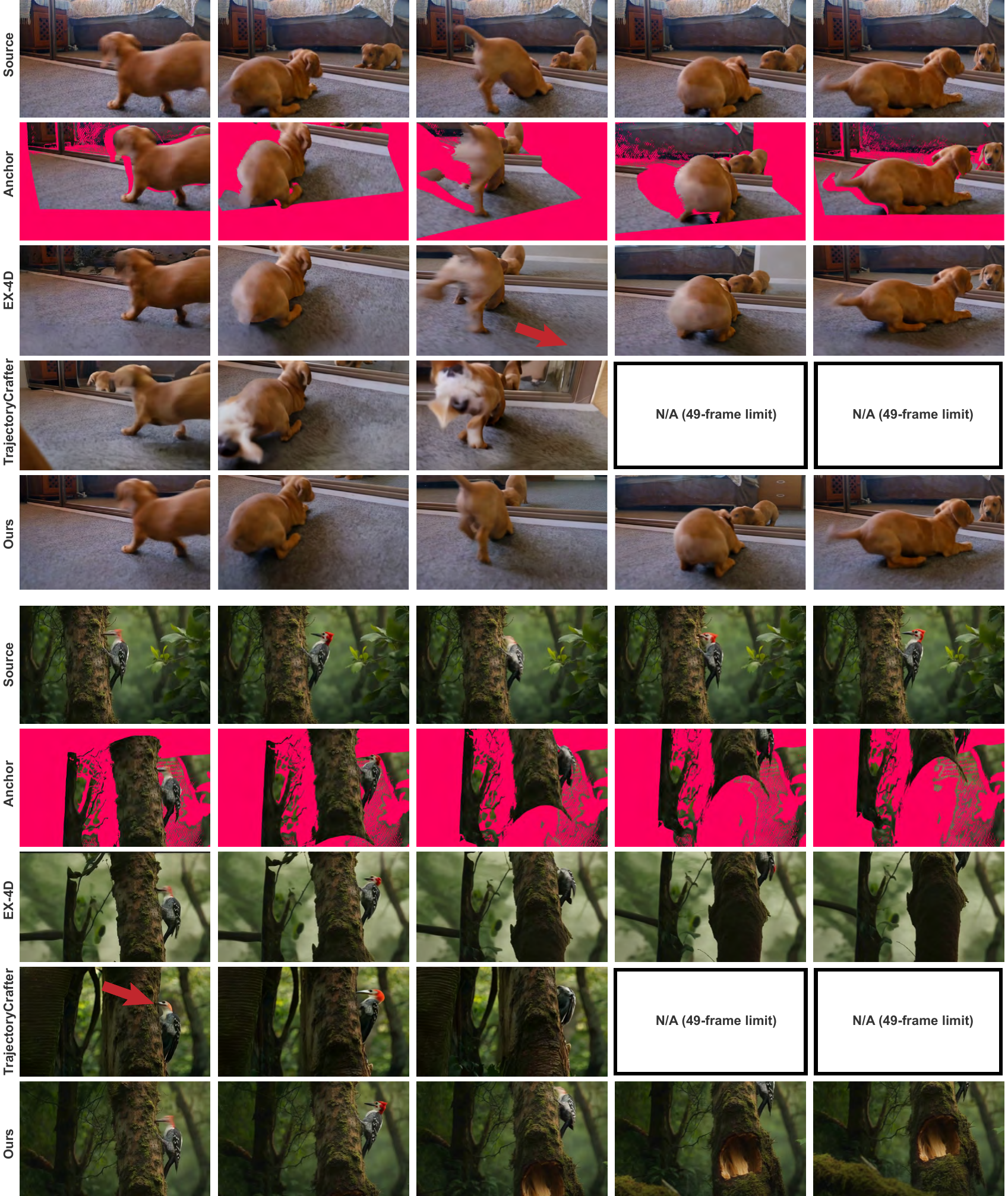}
    \vspace{-0.25in}
\caption{\textbf{Handling Random Camera Start Points while Maintaining Detail.} We illustrate anchor trajectories initiated from arbitrary viewpoints, causing significant spatial misalignment at the start of generation. Our model successfully handles these large initial viewpoint shifts without compromising quality, consistently preserving fine details and original textures.}
    \label{fig:supp_comparison5}
    \vspace{-0.2in}
\end{figure*}

\begin{figure*}
    \centering
    \includegraphics[width=\linewidth]{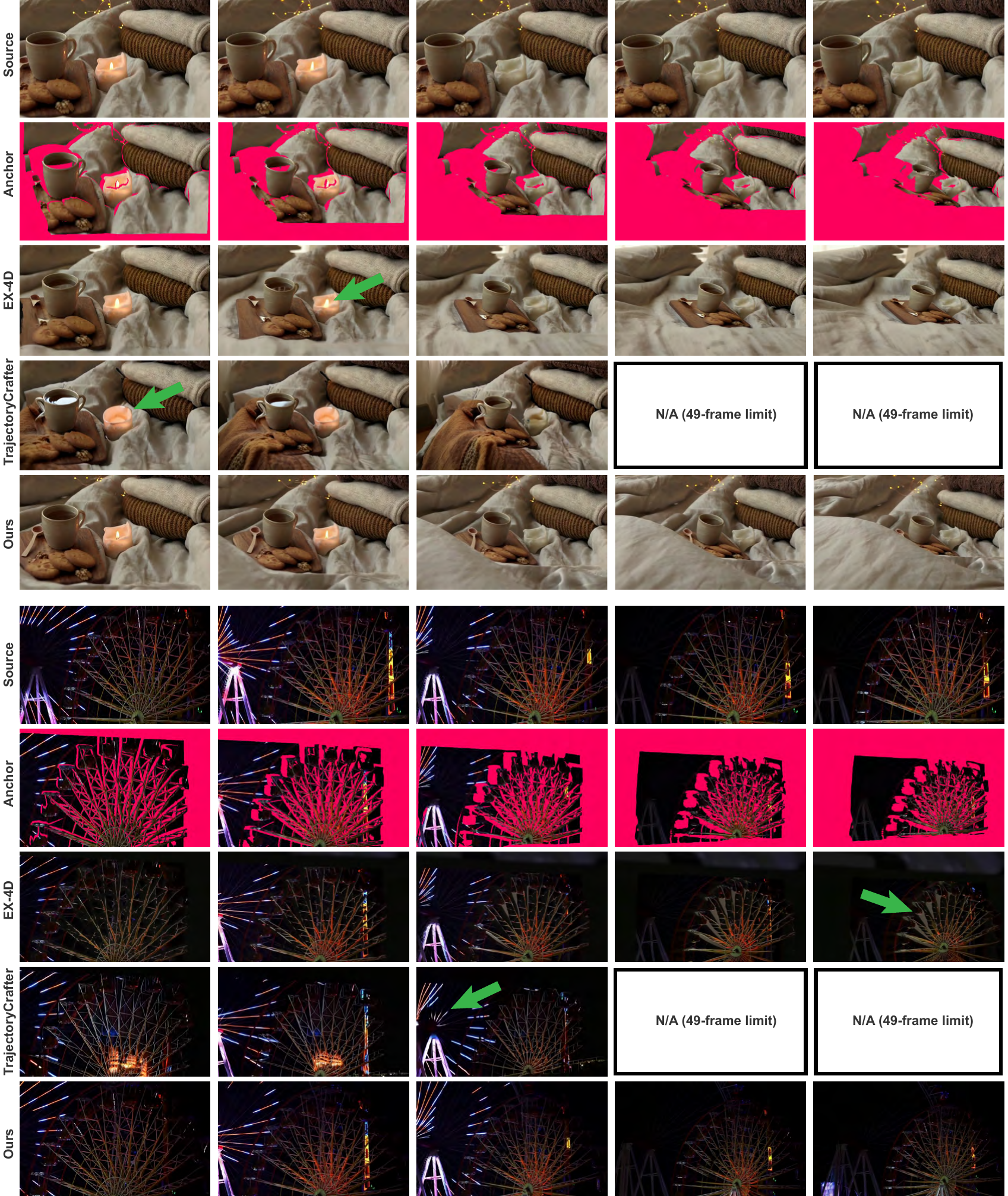}
    \vspace{-0.25in}
\caption{\textbf{Handling Random Camera Start Points while Maintaining Detail.} Additional examples of trajectories initiated from arbitrary viewpoints. Our model maintains exceptional stability and texture preservation under these challenging initialization conditions.}
    \label{fig:supp_comparison6}
    \vspace{-0.2in}
\end{figure*}

\begin{figure}
    \centering
    \includegraphics[width=\linewidth]{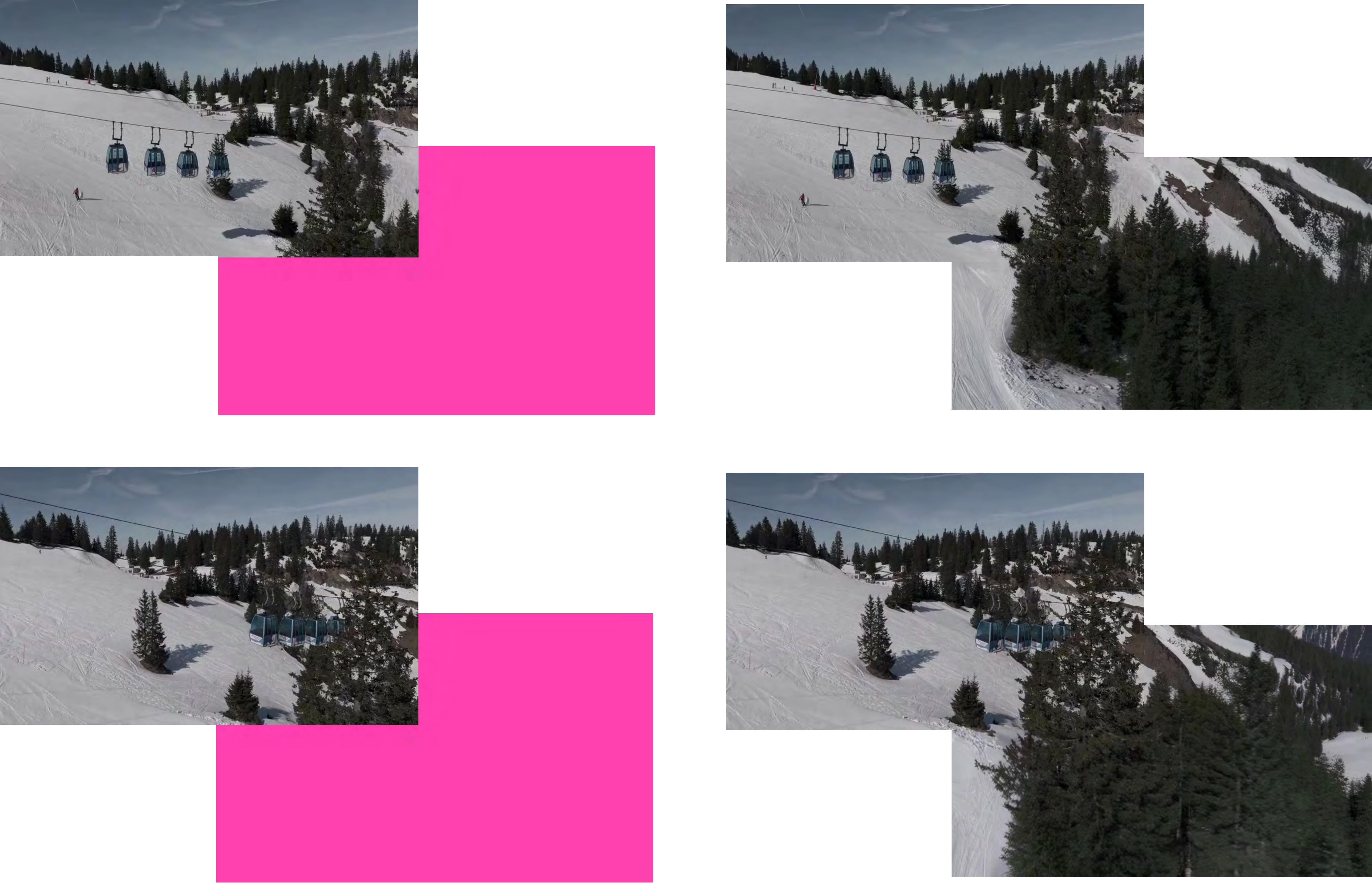}
    \vspace{-0.25in}
\caption{\textbf{Extending Capabilities: Generative Outpainting.} Beyond standard video reshooting, our model demonstrates strong generative priors useful for outpainting tasks. In this example, a 2D crop window is shifted significantly towards the bottom-right, revealing a large unseen region. Crucially, unlike standard per-frame outpainting, our approach is full-context. The model attends to the entire input source video simultaneously, leveraging its learned implicit routing to plausibly hallucinate missing content and coherently complete the scene with high temporal stability.}
    \label{fig:supp_crop}
    \vspace{-0.2in}
\end{figure}

In this supplementary material, we provide comprehensive implementation details, extended experimental results, and a discussion of potential applications for our video reshooting framework.

\section{Motivation} 
\label{sec:supp_motivation}

Achieving photorealistic video reshooting requires simultaneously maintaining precise geometric alignment with a novel camera trajectory while preserving the intricate textures and dynamic content of the original source video. Existing approaches often struggle to balance these requirements, leading to characteristic failure modes illustrated in Figure~\ref{fig:supp_motivation}. Methods that rely solely on sparse or imperfect anchor videos for conditioning, such as \exfd{}~\cite{ex4d}, are forced to hallucinate missing texture information without ground truth, frequently resulting in deformed textures or severe ghosting artifacts. Conversely, models trained primarily on synthetic datasets, like \recammaster{}~\cite{bai2025recammaster}, often exhibit a domain gap when applied to real-world footage, failing to capture complex physical dynamics and fine-grained details. These limitations motivate our approach, which utilizes a self-supervised framework designed to learn directly from real-world monocular videos and explicitly fuses high-fidelity source textures with anchor-based geometric guidance.

\section{Implementation Details}
\subsection{Diffusion Transformer Configuration}
\label{sec:diffusion_transformer_config}

Our diffusion transformer is built upon the \textbf{Wan2.2-I2V 14B} model~\cite{wan21}. This base model employs a Mixture-of-Experts (MoE) design, featuring distinct parameter sets specialized for high-SNR (low-noise) and low-SNR (high-noise) regions of the diffusion trajectory. The architecture functions as a Latent Diffusion Model (LDM), performing all video generation within a compressed latent space. Videos are encoded into a latent representation $z \in \mathbb{R}^{C_L \times T_L \times H_L \times W_L}$ by a causal 3D VAE with a U-Net backbone. This VAE temporally compresses the input, mapping the input video frames to $T_L$ latent frames.

The native Wan2.2-I2V model is designed to condition on an image-reference latent, along with a binary mask indicating visible frame locations. Its standard conditioning pathway concatenates this image-reference latent, a $C_M$-channel binary mask, and a $C_L$-channel noisy latent along the channel dimension. This results in a tensor of shape $\mathbb{R}^{(2C_L + C_M) \times T_L \times H_L \times W_L}$, which is then patchified and processed by the DiT blocks utilizing 3D Rotary Positional Embeddings (3D-RoPE).

For our video reshooting task, we adapt this scheme to integrate both anchor and source video information:
\begin{description}[noitemsep, topsep=0pt, partopsep=0pt, leftmargin=1em, style=unboxed, font=\normalfont\bfseries]
    \item[Anchor Conditioning Stream] The VAE-encoded anchor latent $z_a$ is concatenated with a $C_L$-channel noise latent $z_n$ and a $C_M$-channel downsampled binary mask $M_a$. This forms a tensor representing the anchor conditional input.
    \item[Source Conditioning Stream] The VAE-encoded source latent $z_s$ is duplicated along the channel dimension (replacing $z_n$) and then concatenated with an all-ones $C_M$-channel mask $M_s$. 
\end{description}
These two conditional inputs are temporally concatenated, leading to a token sequence that is twice as long as standard inference (i.e., $2 \times T_L$ latent frames). We share the parameters of the patchify blocks across both pathways.

To distinguish the positional context of the appended source tokens, we apply a constant offset to their 3D-RoPE along the temporal dimension within the DiT blocks. This offset magnitude is set to 50, which significantly exceeds our maximum number of training latent frames ($T_L = 20$). This decoupling allows us to flexibly change the length of generated videos during inference. The self-attention layers within the DiT blocks attend jointly over all concatenated anchor and source tokens, enabling efficient cross-video knowledge transfer. Finally, we apply rank-512 LoRA fine-tuning to all attention and fully connected layers.

\subsection{Augmentations and Ablation Setup}
\label{sec:augmentations_ablation_setup}

As introduced in the main paper, we implement several technical augmentations designed to enhance model robustness and visual quality.

\begin{description}[noitemsep, topsep=0pt, partopsep=0pt, leftmargin=0pt, style=unboxed, font=\normalfont\bfseries]
    \item[Auxiliary Loss] To ensure the source token pathway actively retains meaningful content, we apply an L1 reconstruction loss between the output tokens corresponding to the source video and the VAE-encoded clean source latent $z_s$. This loss is weighted by a factor of 0.1.
    
    \item[Fluorescent Background Anchor] In standard anchor video generation, regions representing new viewpoints or disocclusions are filled with a black background. However, for dark scenes, the model struggles to distinguish masked regions from actual dark content. We found that replacing the standard black background with a high-contrast fluorescent pink color provides a distinct signal for the model, making the boundary of missing information unambiguous.

    \item[Random Query] Our default anchor generation warps the first frame of the source video ($V_s[0]$) to create $V_a$. However, to make the training process more robust to diverse tracking conditions, we introduce an augmentation where the reference frame for dense tracking and warping is randomly selected within the source video ($V_s[t]$). This prevents the model from developing a bias towards early frames and encourages sustained attention to geometric guidance throughout the sequence.

    \item[3D Noise in Anchor] A potential issue arises when the synthesized anchor video ($V_a$) closely resembles the target video ($V_t$), tempting the model to directly copy texture from $V_a$ instead of routing from $V_s$. To suppress $V_a$'s texture information while preserving its 3D geometric guidance, we inject Gaussian noise into the RGB values of the reference frame ($V_s[t]$) \textit{before} it is forward-warped. The noise magnitude is sampled uniformly between [0, 0.5] per channel. Unlike injecting noise \textit{after} warping, pre-warping noise ensures that the noise itself moves coherently with the underlying 3D structure. This forces the model to rely on $V_s$ for high-fidelity content while still inferring motion from $V_a$.
\end{description}

\subsection{Anchor Generation for Inference}
\label{sec:inference_anchor_gen}

While our self-supervised training pipeline utilizes efficient 2D warping, during inference we utilize an explicit 3D projection approach to define the target camera trajectory. We extract dense geometric information from the source video using state-of-the-art monocular depth and camera estimation models~\cite{huang2025vipe, xu2025geometrycrafter, 4d_recon_hu2025depthcrafter}. Using these depth maps and camera parameters, we unproject the source video pixels into 3D world space, resulting in a consistent, colored point cloud. Finally, to generate an anchor frame corresponding to a target camera pose, we re-render this 3D point cloud from the novel viewpoint, utilizing the input cameras to cancel out the original camera motion.

\subsection{Evaluation}

\textbf{Dataset.} We constructed a final set of 100 videos, sub-sampled from 1,000 stratified examples from the Opensora-mixkit dataset~\cite{lin2024open}. Videos were randomly assigned 1 of 10 predefined camera motion trajectories. To ensure balanced movement representation, we calculated the mean value of the generated anchor mask ($M_a$) for each video, grouped them by trajectory, and selected the 10 videos nearest to their respective group median. This mitigates outliers with extreme mask presence while preserving the underlying semantic distribution.

\textbf{Metrics.} We evaluate along three key dimensions:
\begin{itemize}[noitemsep, topsep=0pt]
    \item \textit{Camera Accuracy:} We measure Rotational Error (RotErr) and Translational Error (TransErr) following~\cite{camera_traj_he2024cameractrl}. We extract camera poses using~\cite{huang2025vipe}, align the first frame, and calculate relative poses between the generated and ground-truth trajectories. TransErr is the sum of squared L2 distances, and RotErr is the sum of angular differences.
    \item \textit{View Synchronization:} To assess perceptual similarity, we calculate the Fréchet Video Distance (FVD-V)~\cite{eval_unterthiner2019fvd} between the source videos and generated outputs. For semantic synchronization, we use CLIP-V, the average frame-wise CLIP cosine similarity at matching timestamps. For fine-grained spatial alignment, we use Matching Pixels (Mat. Pix) via the GIM model~\cite{eval_matpix}, counting geometrically-verified (RANSAC) feature matches that exceed a confidence threshold of 0.5.
    \item \textit{Video Quality:} We utilize the VBench benchmark~\cite{eval_huang2024vbench} for perceptual quality. To evaluate temporal consistency, we compute CLIP-F, the average CLIP similarity score of adjacent generated frames.
\end{itemize}

\subsection{Extended Qualitative Comparisons and Applications} 
\label{sec:supp_comparisons}

To provide a comprehensive visual analysis, we present extended qualitative comparisons in Figures~\ref{fig:supp_comparison1}, \ref{fig:supp_comparison2}, and \ref{fig:supp_comparison3}. Baseline methods frequently exhibit characteristic artifacts, such as severe blurring, geometric distortion, or the loss of fine-grained details present in the source input. In contrast, our approach consistently demonstrates superior visual fidelity by effectively leveraging the stable scene priors provided by the source-video conditioning.

Furthermore, Figures~\ref{fig:supp_comparison5} and \ref{fig:supp_comparison6} demonstrate the robustness of our model to challenging initialization conditions, enabled by our Random Query augmentation. Even when target trajectories initiate from random viewpoints causing massive spatial misalignment at the first frame, our model exhibits remarkable stability.

Finally, we highlight the versatility of our generative framework. Beyond standard reshooting, Figure~\ref{fig:supp_crop} illustrates the model's capacity for generative outpainting. When subjected to extreme 2D cropping that reveals previously unseen regions, the model does not simply hallucinate content frame-by-frame. Instead, it performs full-context outpainting. By attending to the entire source video sequence simultaneously via our conditioning architecture, the model can route available texture information from other temporal frames to plausibly and coherently fill the missing regions, ensuring strict temporal consistency.

\end{document}